\documentclass[a4paper]{svproc}
\usepackage{amsmath} 
\usepackage{amssymb}  
\usepackage{bm}  
\usepackage{graphics} 
\usepackage{epsfig} 
\usepackage{wrapfig}
\usepackage{subcaption}
\usepackage[font=small,labelfont=small]{caption}
\usepackage{hyperref}
\usepackage{color}
\usepackage[table]{xcolor}
\usepackage[ruled, linesnumbered]{algorithm2e}
\usepackage{tikz}
\usetikzlibrary{arrows.meta}

\usepackage{url}

\newcommand{\smalleq}[1]{\scalebox{0.95}{#1}}

\newcommand{\euclideanspace}{\mathbb{R}}
\newcommand{\manifold}{\mathcal{M}}
\newcommand{\configmanifold}{\mathcal{Q}}
\newcommand{\tangentspace}[1]{\mathcal{T}_{#1}\mathcal{M}}
\newcommand{\configtangentspace}[1]{\mathcal{T}_{#1}\mathcal{Q}}

\newcommand{\innerprod}[3]{\langle #2, #3 \rangle_{#1}}  
\newcommand{\jointposition}{\bm{q}}
\newcommand{\jointvelocity}{\dot{\bm{q}}}
\newcommand{\jointacceleration}{\ddot{\bm{q}}}

\definecolor{darkyellow}{rgb}{0.86, 0.66, 0.}
\definecolor{darkred}{rgb}{0.66, 0.08, 0.23}
\definecolor{darkorange}{rgb}{1., 0.75, 0.0}
\definecolor{orange}{rgb}{1., 0.65, 0.1}
\definecolor{skyblue}{rgb}{0., 0.65, 0.9}
\definecolor{steelblue}{rgb}{0.275, 0.501, 0.706}
\definecolor{steelblue2}{rgb}{0.3, 0.4, 0.45}
\definecolor{gray}{rgb}{0.7, 0.7, 0.7}
\definecolor{darkblue}{rgb}{0, 0.2, 0.745}
\definecolor{purple}{rgb}{0.7, 0.296, 0.9}

\DeclareRobustCommand{\bluegeodesic}{\raisebox{2pt}{\tikz{\draw[steelblue2,solid,line width = 1.5pt](0,0) -- (3mm,0);}}}
\DeclareRobustCommand{\bluestraight}{\raisebox{2pt}{\tikz{\draw[steelblue2,densely dashed,line width = 1.5pt](0,0) -- (3mm,0);}}}
\DeclareRobustCommand{\blackmetric}{\tikz{ \filldraw[color=black, fill=white!60, thick](0,0) ellipse (.15 and 0.08);}}
\DeclareRobustCommand{\robotfirstGS}{\raisebox{2pt}{\tikz{\draw[darkorange,solid,line width = 2.5pt](0,0) -- (3mm,0);}}}
\DeclareRobustCommand{\robotsecondGS}{\raisebox{2pt}{\tikz{\draw[darkblue,solid,line width = 2.5pt](0,0) -- (3mm,0);}}}
\DeclareRobustCommand{\robotGScomb}{\raisebox{2pt}{\tikz{\draw[purple,solid,line width = 2.5pt](0,0) -- (3mm,0);}}}

\usepackage[per-mode=fraction]{siunitx}
\usepackage{comment}
\usepackage[normalem]{ulem}
\usepackage{xspace}
\usepackage{xcolor}

\DeclareSIUnit\px{px}
\DeclareSIUnit\fps{fps}

\definecolor{OliveGreen}{RGB}{0,200,25}
\newcommand{\red}[1]{\textcolor{red}{#1}}

\newcommand{\darkgreen}[1]{\textcolor{OliveGreen}{#1}}

\newif\iffinal

\newcommand{\replaced}[2]{%
	\iffinal%
	#2%
	\else%
	\red{\ifmmode\text{\sout{\ensuremath{#1}}}\else\sout{#1}\fi}\darkgreen{#2}%
	\fi%
}
\newcommand{\removed}[1]{%
	\iffinal%
	\else%
	\red{\ifmmode\text{\sout{\ensuremath{#1}}}\else\sout{#1}\fi}%
	\fi%
}

\graphicspath{{Figures/}}

\begin{document}
\mainmatter              
\title{Riemannian geometry as a unifying theory \\for robot motion learning and control}
\titlerunning{Riemannian geometry as a unifying theory for robot motion}  
%
\author{No\'emie Jaquier \and Tamim Asfour}
\authorrunning{N. Jaquier and T. Asfour} 
%
\tocauthor{No\'emie Jaquier and Tamim Asfour}
\institute{Karlsruhe Institute of Technology, Institute for Anthropomatics and Robotics\\ 76131 Karlsruhe, Germany, 
\email{\{noemie.jaquier, asfour\}@kit.edu}}

\maketitle              

\vspace{-0.2cm}
\begin{abstract}
Riemannian geometry is a mathematical field which has been the cornerstone of revolutionary scientific discoveries such as the theory of general relativity. 
Despite early uses in robot design and recent applications for exploiting data with specific geometries, it mostly remains overlooked in robotics. With this blue sky paper, we argue that Riemannian geometry provides the most suitable tools to analyze and generate well-coordinated, energy-efficient motions of robots with many degrees of freedom. 
Via preliminary solutions and novel research directions, we discuss how Riemannian geometry may be leveraged to design and combine physically-meaningful synergies for robotics, and how this theory also opens the door to coupling motion synergies with perceptual inputs. 

\end{abstract}

\section{Motivation}
\label{Sec:Motivation}
The last years have seen the emergence of various, complex robotics systems with increased number of degrees-of-freedom (DoF). These include humanoid and quadrupedal robots, exoskeletons and robotic hands, among others. Despite recent progress, these robots remain to be actively deployed in our everyday life. Among the challenges that remain unsolved is the generation of well-coordinated, energy-efficient, and reliable robot motions under any circumstances. This problem is further exacerbated by high number of DoF.

As opposed to robots, humans are naturally able to generate skillful motions. For instance, they efficiently plan optimal trajectories while coping with the redundancy of their high-dimensional configuration space. Moreover, these dexterous motions usually adapt to the perceived environment. 
Following insights from neurosciences, roboticists have designed biologically-inspired solutions to cope with the redundancy of robotic systems. In particular, the notion of muscle or movement \emph{synergies} --- coherent co-activation of motor signals~\cite{dAvella03:MuscleSynergies} --- has been widely used to generate motions of highly-redundant robots such as robotic hands~\cite{Gabiccini11:HandSynergiesGraspingForces,Santello98:PosturalHandSynergies} and humanoid robots~\cite{Gu06:HumanoidSynergies,Hauser07:KinematicSynergiesBalanceControl}. Synergies offer an elegant and powerful alternative to classical control schemes, as a wide range of motions is simply generated by combining few well-selected canonical synergies. 
However, synergies have often been extracted using linear methods, e.g., principal component analysis (PCA), which disregard the nonlinear nature of human and robot configuration spaces. Latent variable models (LVMs) based on Gaussian processes~\cite{Romero13:PosturalSynergiesRoboticGrasping} or auto-encoders~\cite{Starke20:ObjectSpecificGraspSynergies} have recently been used to cope with this nonlinearity. However, the resulting 
synergy spaces are physically meaningless and hard to interpret, thus limiting the applicability of these methods. Moreover, although synergies have been recently adapted to object properties in robot grasping~\cite{Starke20:ObjectSpecificGraspSynergies}, linking synergies with perceptual inputs still remains an open problem.

Instead, the nonlinear nature of mechanical systems such as humans and robots is suitably described by Riemannian geometry.
Indeed, the configuration space of any multi-linked mechanical system can be identified with a Riemannian manifold, i.e., a smooth curved geometric space incorporating the structural and inertial
characteristics of the system~\cite{Bullo05:GeometricControl}. On this basis, Riemannian computational models of human motion planning were recently proposed~\cite{Biess07:ComputationalModelPointing,Klein22:SequenceGeodesicSynergies,Neilson15:GeodesicSynergyHypothesis}. These models aim at uncovering the optimization principles underlying human motion generation, and thus the key mechanisms coping with the redundancy of the human body. They suggest that human motions are planned as \emph{geodesics}, i.e., minimum-muscular-effort (or shortest) paths, in the configuration space manifold. As shown by Neilson et al.~\cite{Neilson15:GeodesicSynergyHypothesis}, each geodesic constitutes \emph{a coherent coordination of joint angles with minimum energy demand}, and thus corresponds to a \emph{geodesic synergy}. In that sense, human point-to-point motions follow a single geodesic synergy~\cite{Biess07:ComputationalModelPointing}, while complex movements may result from sequences~\cite{Klein22:SequenceGeodesicSynergies} or compositions~\cite{Neilson15:GeodesicSynergyHypothesis} of these.
Despite the soundness of Riemannian geometry to describe configuration spaces and the attractive properties of geodesic synergies, it remains overlooked by the robotics community. 

In this paper, we contend that Riemannian geometry offers the most suitable mathematical tools to \emph{(i)} extract interpretable nonlinear movement synergies for robotics, \emph{(ii)} combine them to generate dexterous robot motions, and \emph{(iii)} intertwine them with various perceptual inputs. Inspired by insights from human motion analysis~\cite{Biess07:ComputationalModelPointing,Klein22:SequenceGeodesicSynergies,Neilson15:GeodesicSynergyHypothesis}, we propose preliminary solutions that exploit Riemannian geometry for the design of meaningful synergies which account for the structural and inertial properties of high-DoF robots. We then describe our envisioned line of research, as well as the important challenges to be tackled. Finally, we discuss how geodesic synergies have the potential to contribute to the deployment of robots in our everyday life not only by reducing their energy requirements, but also by generating natural and interpretable motions of humanoid and wearable robots, which adapt to perceptual inputs. 

It is worth noticing that geometric methods have been successfully applied to robotics from early on, e.g., for robot design~\cite{Park95:RobotDesign}, or for formulating robot kinematics and dynamics models~\cite{Selig05}. Additionally, Riemannian methods recently gained interest in robot learning and control to handle data with particular geometries~\cite{Jaquier21:ManipLearningTracking,Jaquier21:GaBOMatern}, capture relevant demonstration patterns~\cite{BeikMohammadi21:GeodesicMotionSkills}, or combine multiple simple policies~\cite{Ratliff18:RiemannianMotionPolicies}. These works exploit Riemannian geometry at data level to learn or define task-specific policies, and thus are complementary to the ideas presented hereafter. In contrast, this paper focuses on the potential of human-inspired, physically-meaningful geometric representations of low-level robot actions (i.e., joint coordinations), and their coupling to perception. 

\section{Geodesic synergies: Basics and Proof of concept}
\label{Sec:GeodesicSynergies}
\vspace{-0.2cm}
In this section, we first briefly introduce the mathematical tools underlying the notion and use of geodesic synergies. For in-depth introductions to Riemannian geometry and geometry of mechanical systems, we refer the interested reader to, e.g.,~\cite{DoCarmo92:RiemannianGeometry,Lee18:RiemannianManifolds}, and to~\cite{Bullo05:GeometricControl}, respectively. We then provide examples of geodesic synergies and preliminary solutions to use them for robot motion generation. 

\vspace{-0.3cm}
\subsection{Riemannian geometry of mechanical systems}
\label{subSec:GeometryMechSys}
\vspace{-0.1cm}
\subsubsection{Riemannian manifolds}
The configuration space $\configmanifold$ of a multi-linked mechanical system can be viewed as a Riemannian manifold, a mathematical space which inherently considers the characteristics of the system. 
A $n$-dimensional manifold $\manifold$ is a topological space which is locally Euclidean. In other words, each point in $\manifold$ has a neighborhood which is homeomorphic to an open subset of the $n$-dimensional Euclidean space $\euclideanspace^n$. 
A tangent space $\tangentspace{\bm{x}}$ is associated to each point $\bm{x}\in\manifold$ and is formed by the differentials at $\bm{x}$ to all curves on $\mathcal{M}$ passing through $\bm{x}$. 
The disjoint union of all tangent spaces $\tangentspace{\bm{x}}$ forms the tangent bundle $\tangentspace{}$.
A Riemannian manifold is a smooth manifold equipped with a Riemannian metric, i.e., a smoothly-varying inner product acting on $\tangentspace{}$. Given a choice of local coordinates, the Riemannian metric is represented as a symmetric positive-definite matrix $\bm{G}(\bm{x})$, called a metric tensor, which depends smoothly on $\bm{x}\in\manifold$. 
The Riemannian metric leads to local, nonlinear expressions of inner products and thus of norms and angles. Specifically, the Riemannian inner product between two velocity vectors $\bm{u}$, $\bm{v}\in\tangentspace{\bm{x}}$ is given as 
$\innerprod{\bm{x}}{\bm{u}}{\bm{v}} = \langle \bm{u}, \bm{G}(\bm{x})\bm{v} \rangle$.

\vspace{-0.4cm}
\subsubsection{The configuration space manifold}
Points on the configuration space manifold correspond to different joint configurations $\jointposition\in\configmanifold$. The manifold $\configmanifold$ can be endowed with the so-called \emph{kinetic-energy} metric~\cite{Bullo05:GeometricControl}. Specifically, the metric tensor $\bm{G}(\jointposition)$ is equal to the mass-inertia matrix of the system at the configuration $\jointposition\in\configmanifold$. In this sense, the mass-inertia matrix, i.e., the Riemannian metric, curves the space so that the configuration manifold accounts for the nonlinear inertial properties of the system. Figure~\ref{subFig:RobotConfigurationSpace} illustrates the effect of the Riemannian metric, represented by ellipsoids $\bm{G}(\jointposition)$ at different joint configurations $\jointposition\in\configmanifold$, on the configuration space of a 2-DoF planar robot. Intuitively, the kinetic energy $k = \frac{1}{2} \langle \jointvelocity, \bm{G}(\jointposition)\jointvelocity \rangle$ at $\jointposition$ is high for velocities $\jointvelocity\in\tangentspace{\jointposition}$ following the ellipsoid major axis, and low along the minor axis. This implies that, in the absence of external force, trajectories of robot joints are not straight lines as in Euclidean space, but instead follow \emph{geodesics}, i.e., generalization of straight lines to manifolds.

\vspace{-0.3cm}
\subsubsection{Geodesics}
\begin{figure}[tbp]
	\centering
	\begin{subfigure}[b]{0.46\textwidth}
	\centering
		\includegraphics[width=1.1\textwidth]{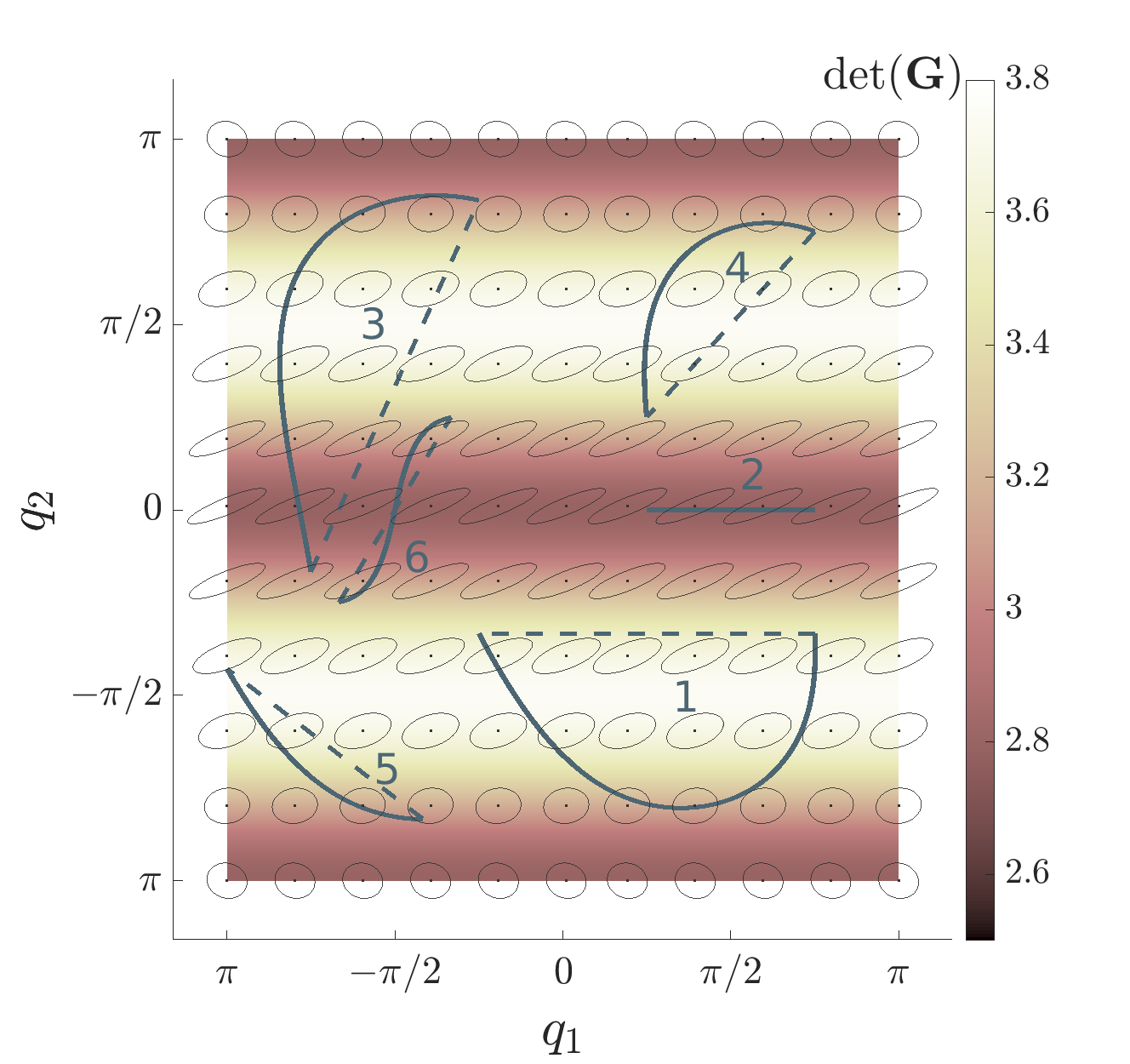}
		\caption{Riemannian configuration space.}
		\label{subFig:RobotConfigurationSpace}
	\end{subfigure}
	\begin{subfigure}[b]{0.26\textwidth}
	\centering
		\includegraphics[width=1.1\textwidth]{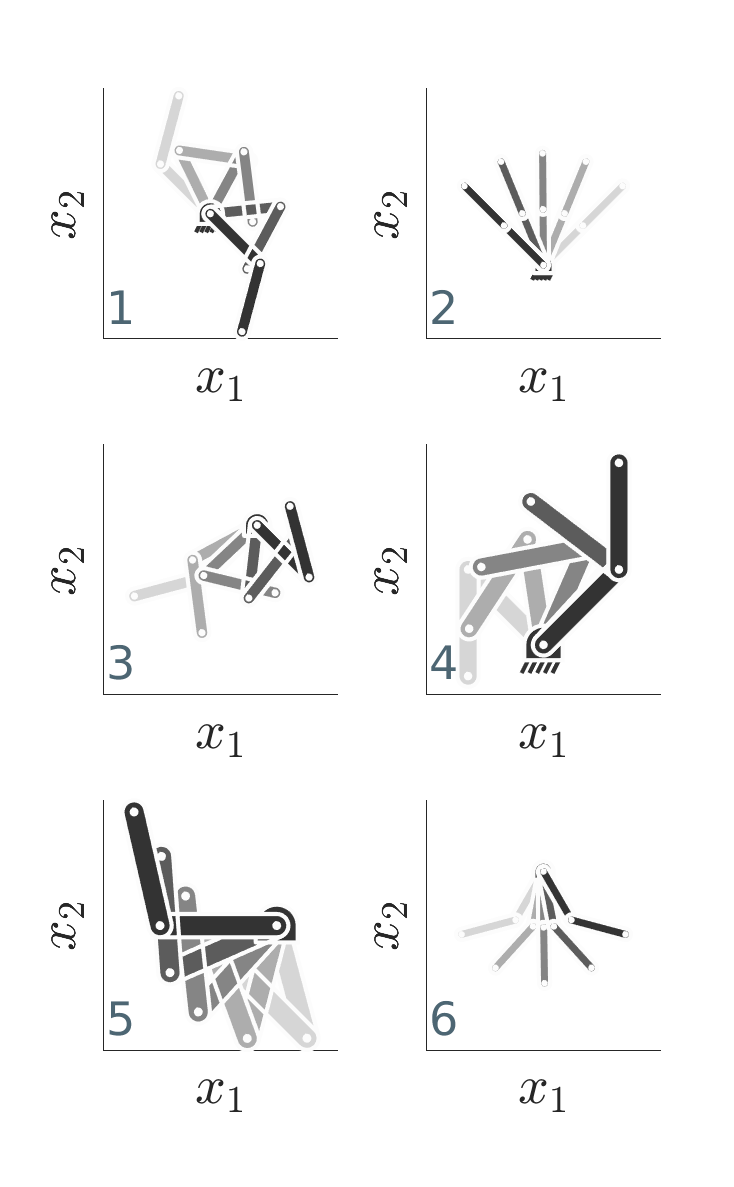}
		\caption{Geodesic motions.}
		\label{subFig:RobotGeodesics}
	\end{subfigure}
	\begin{subfigure}[b]{0.26\textwidth}
	\centering
		\includegraphics[width=1.1\textwidth]{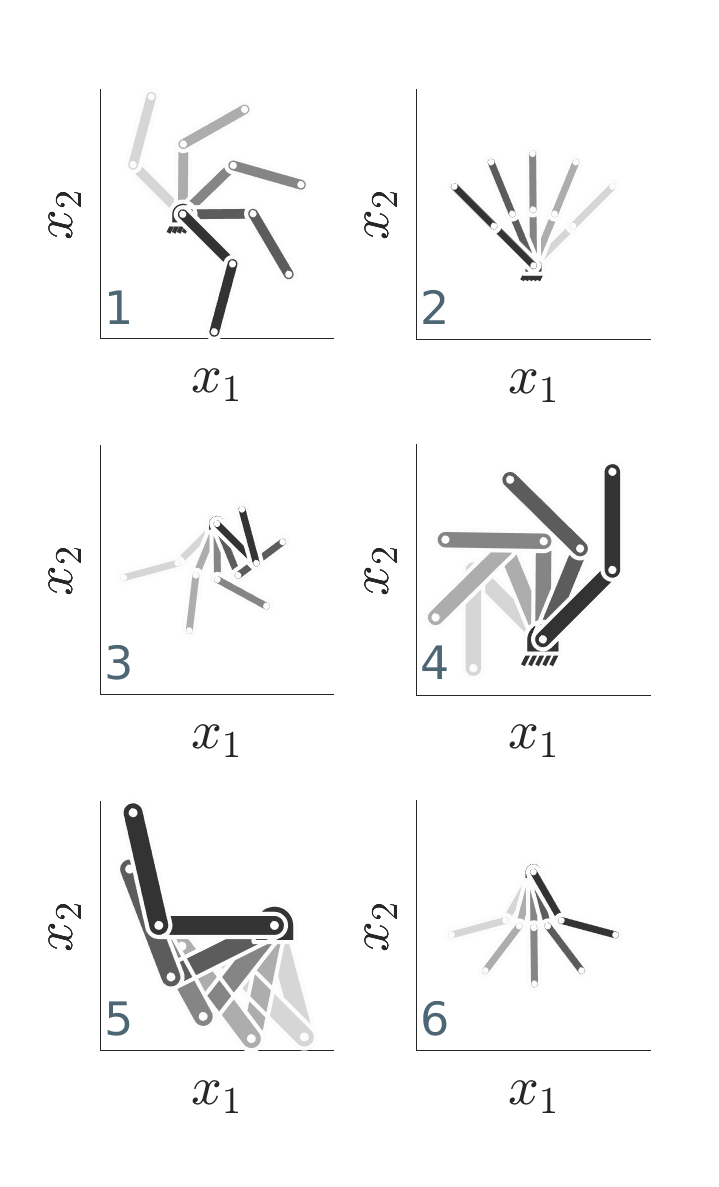}
		\caption{Euclidean motions.}
		\label{subFig:RobotGeodesics}
	\end{subfigure}
	\caption{Illustration of the Riemannian configuration space of a 2-DoFs planar robot. \emph{(a)} The Riemannian metric $\bm{G}(\jointposition)$ (\blackmetric), equal to the robot mass-inertia matrix, curves the space. The energy is reflected by its volume $\propto\text{det}(\bm{G})$. Minimum-energy trajectories between two joint configurations correspond to geodesics (\bluegeodesic), which differ from Euclidean paths (\bluestraight). \emph{(b)-(c)} Robot configurations along the trajectories depicted in \emph{(a)}. The motions reflect the differences between geodesic and Euclidean paths (see 1, 3, 4).}
	\label{Fig:PlanarExample}
	\vspace{-0.3cm}
\end{figure}

Similarly to straight lines in Euclidean space, geodesics are minimum-energy and minimum-length, constant-velocity curves on Riemannian manifolds.
They solve the following system of second-order ordinary differential equations (ODE), obtained by applying the Euler-Lagrange equations to the kinetic energy
\begin{equation}
    \sum\nolimits_{j} g_{ij}(\jointposition) \ddot{q}_j + \sum\nolimits_{jk} \Gamma_{ijk} \dot{q}_j \dot{q}_k = 0,
    \label{Eq:Geodesic}
\end{equation}
where \smalleq{$\sum_{jk} \Gamma_{ijk} \dot{q}_j \dot{q}_k= c_i(\jointposition, \jointvelocity)$} 
represents the influence of Coriolis forces, $g_{ij}$ denotes the $(i,j)$th entry of $\bm{G}$, and \smalleq{$\Gamma_{ijk} = \frac{1}{2} \left( \frac{\partial g_{ij}}{\partial q_k} + \frac{\partial g_{ik}}{\partial q_j} - \frac{\partial g_{jk}}{\partial q_i} \right)$} are the Christoffel symbols of the first kind. 
In other words, geodesic trajectories are obtained by applying the joint acceleration $\jointacceleration(t)$ solution of~\eqref{Eq:Geodesic} at each configuration $\jointposition(t)$ with velocity $\jointvelocity(t)$ along the trajectory. 
Note that~\eqref{Eq:Geodesic} corresponds to the standard equation of motion $\bm{G}(\jointposition)\jointacceleration + \bm{C}(\jointposition,\jointvelocity)\jointvelocity + \bm{\tau}_{\bm{g}}(\jointposition) = \bm{\tau}$ in the absence of gravity ($\bm{\tau}_{\bm{g}}(\jointposition)=0$) and external forces ($\bm{\tau}=0$). Thus, a geodesic corresponds to a passive trajectory of the system.
Geodesic trajectories of a 2-DoF planar robot are displayed in Figure~\ref{Fig:PlanarExample}. As geodesics naturally result in coherent co-activations of joints, they can straightforwardly be used to define synergies, as explained next.

\vspace{-0.3cm}
\subsection{Robot motion generation with geodesic synergies}
\label{subSec:GeodesicMotionGeneration}
\begin{figure}[tbp]
	\centering
	\begin{subfigure}[b]{0.3\textwidth}
		\includegraphics[width=\textwidth,trim={5cm 17cm 10cm 5cm},clip]{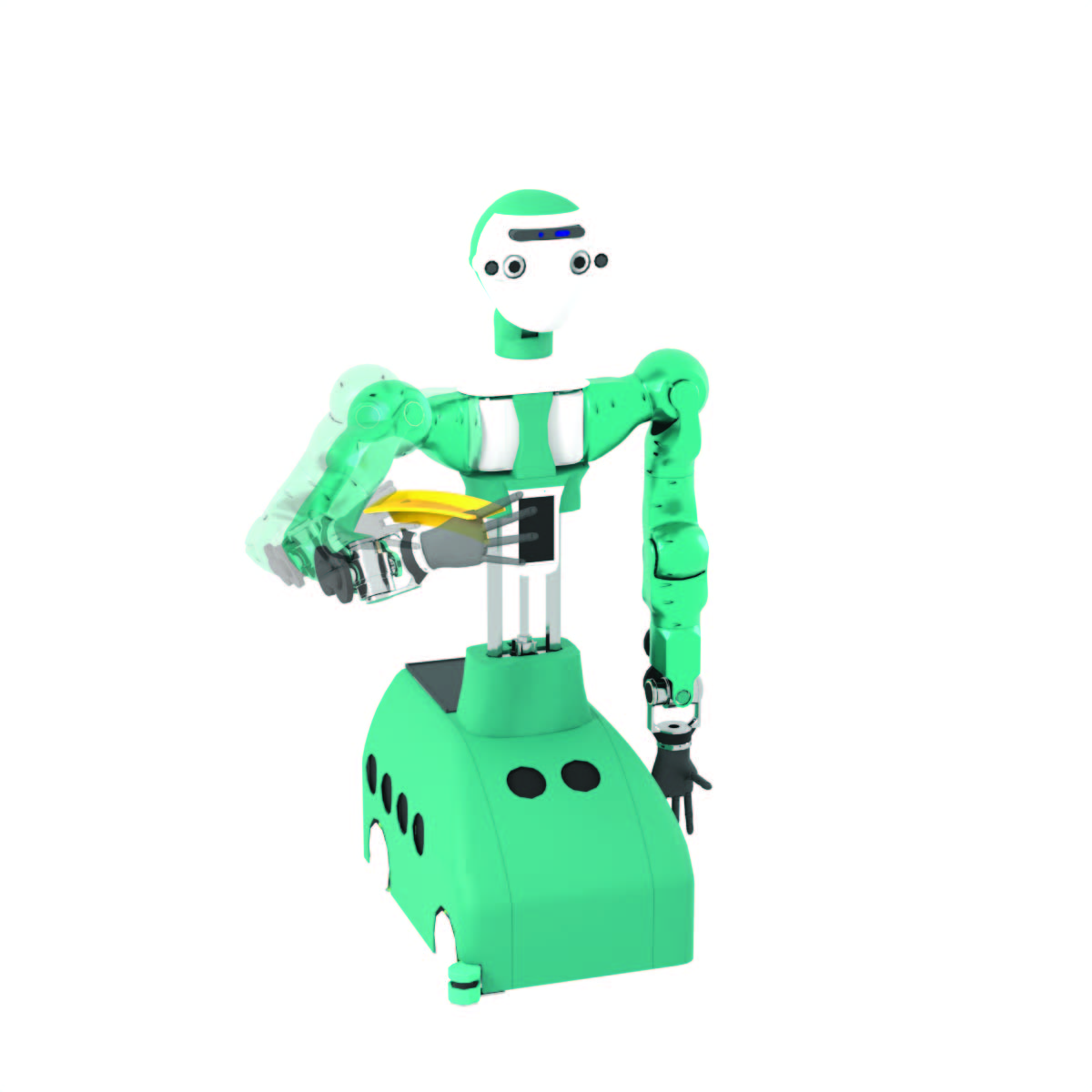}
		\caption{1st geodesic synergy}
		\label{subFig:Geodesic1}
	\end{subfigure}
	\begin{subfigure}[b]{0.3\textwidth}
		\includegraphics[width=\textwidth,trim={5cm 17cm 10cm 5cm},clip]{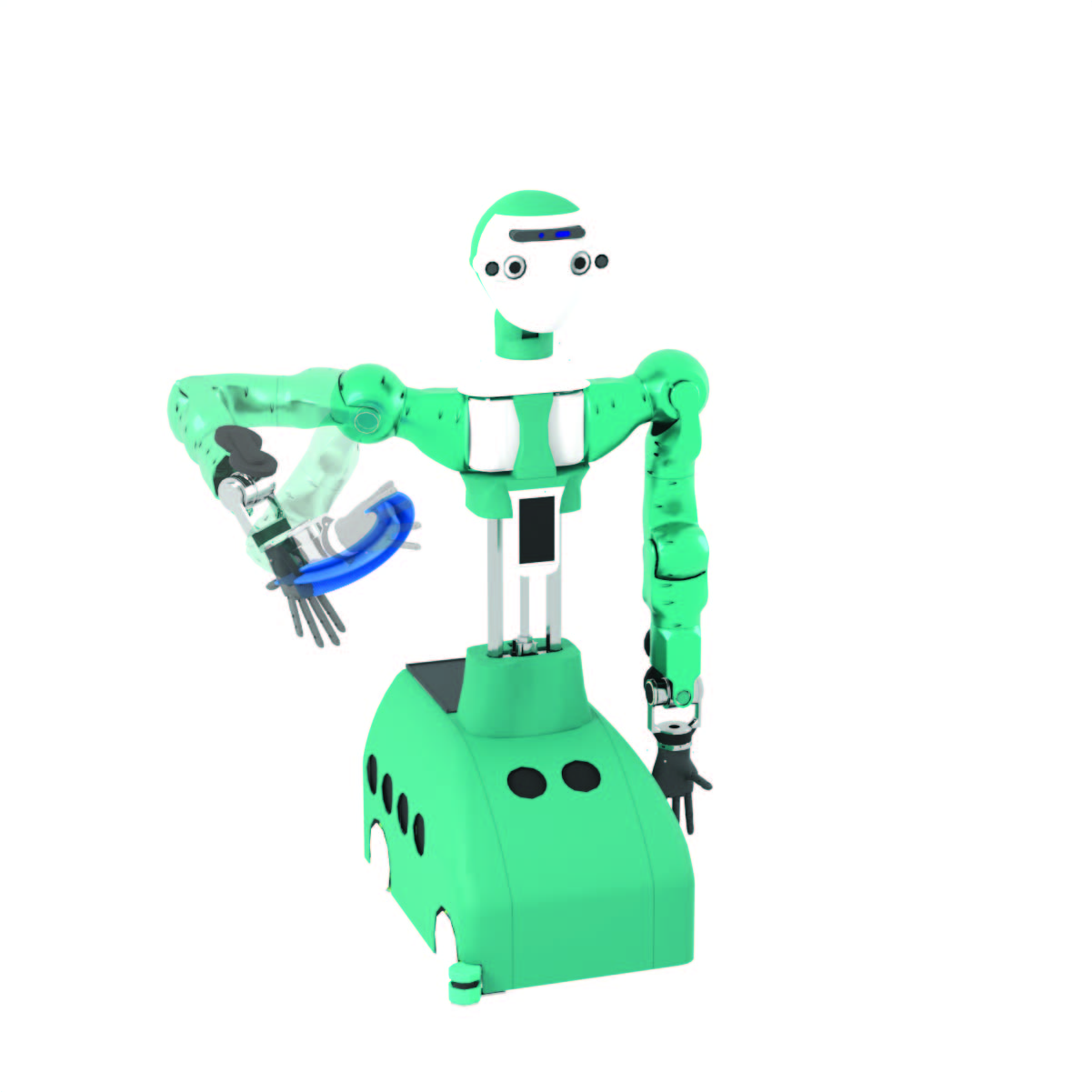}
		\caption{2nd geodesic synergy}
		\label{subFig:Geodesic2}
	\end{subfigure}
	\begin{subfigure}[b]{0.3\textwidth}
		\includegraphics[width=\textwidth,trim={5cm 17cm 10cm 5cm},clip]{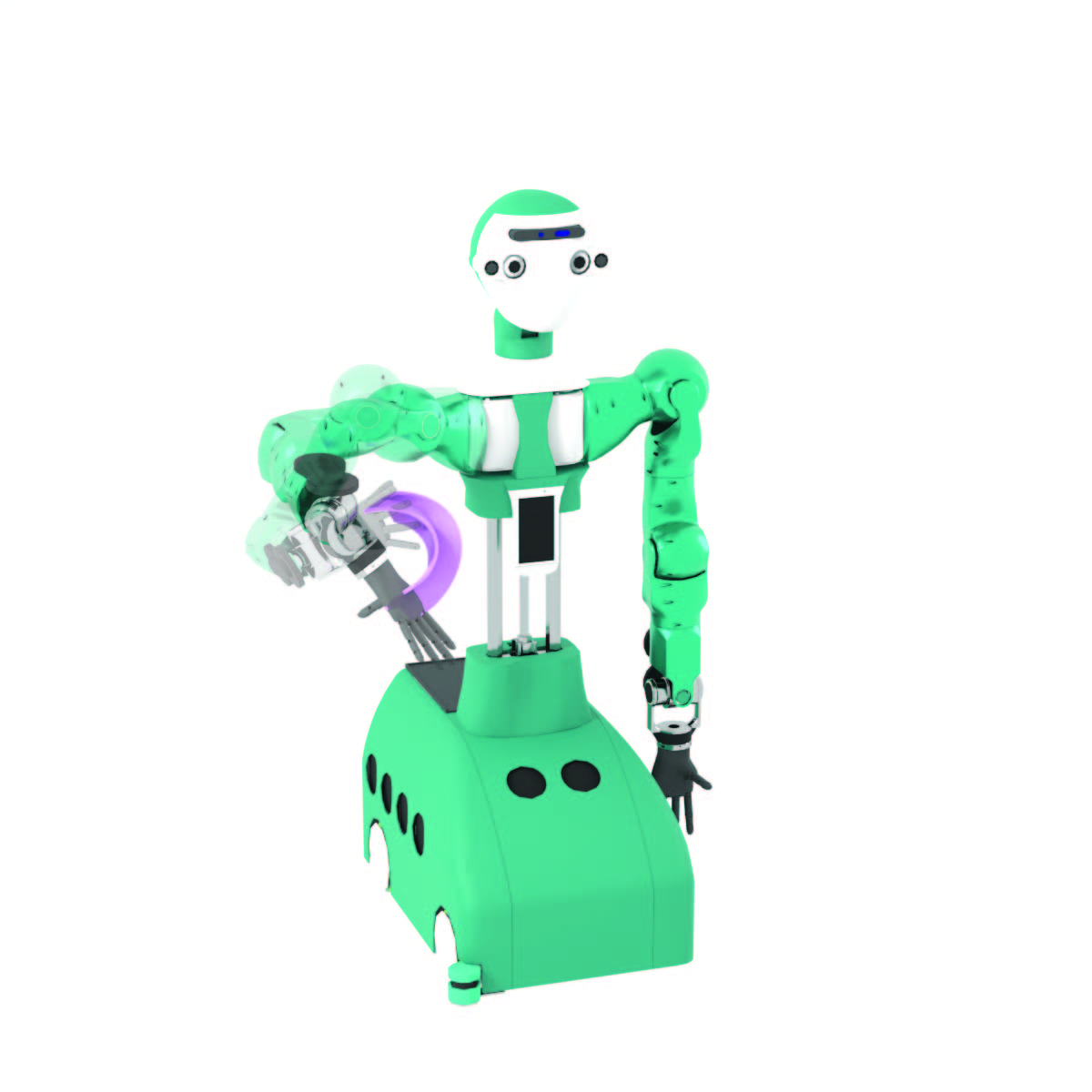}
		\caption{$w_1=1.2$, $w_2=0.7$}
		\label{subFig:Combination1}
	\end{subfigure}
	\begin{subfigure}[b]{0.3\textwidth}
		\includegraphics[width=\textwidth,trim={5cm 17cm 10cm 5cm},clip]{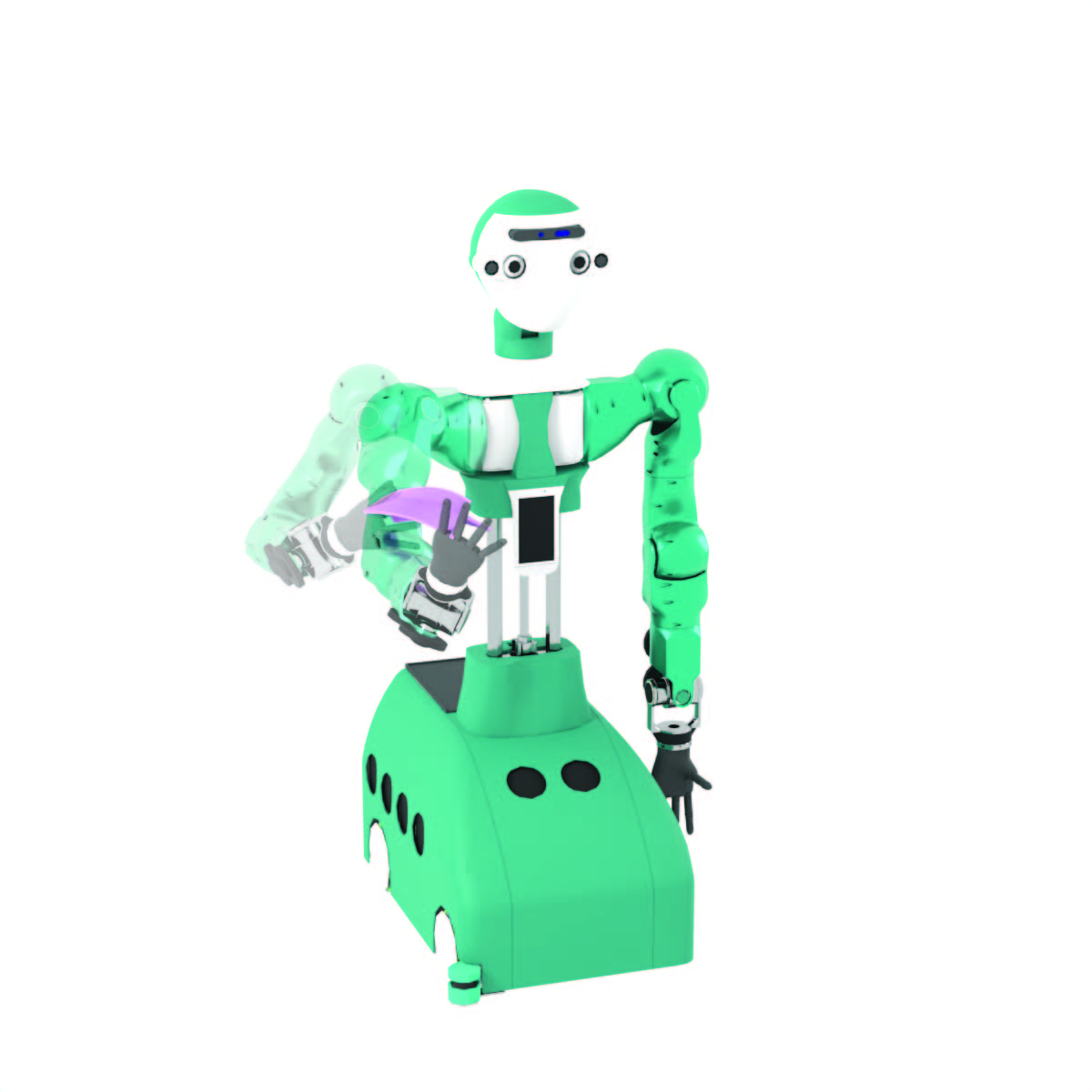}
		\caption{$w_1=1.2$, $w_2=-0.5$}
		\label{subFig:Combination2}
	\end{subfigure}
	\begin{subfigure}[b]{0.3\textwidth}
		\includegraphics[width=\textwidth,trim={5cm 17cm 10cm 5cm},clip]{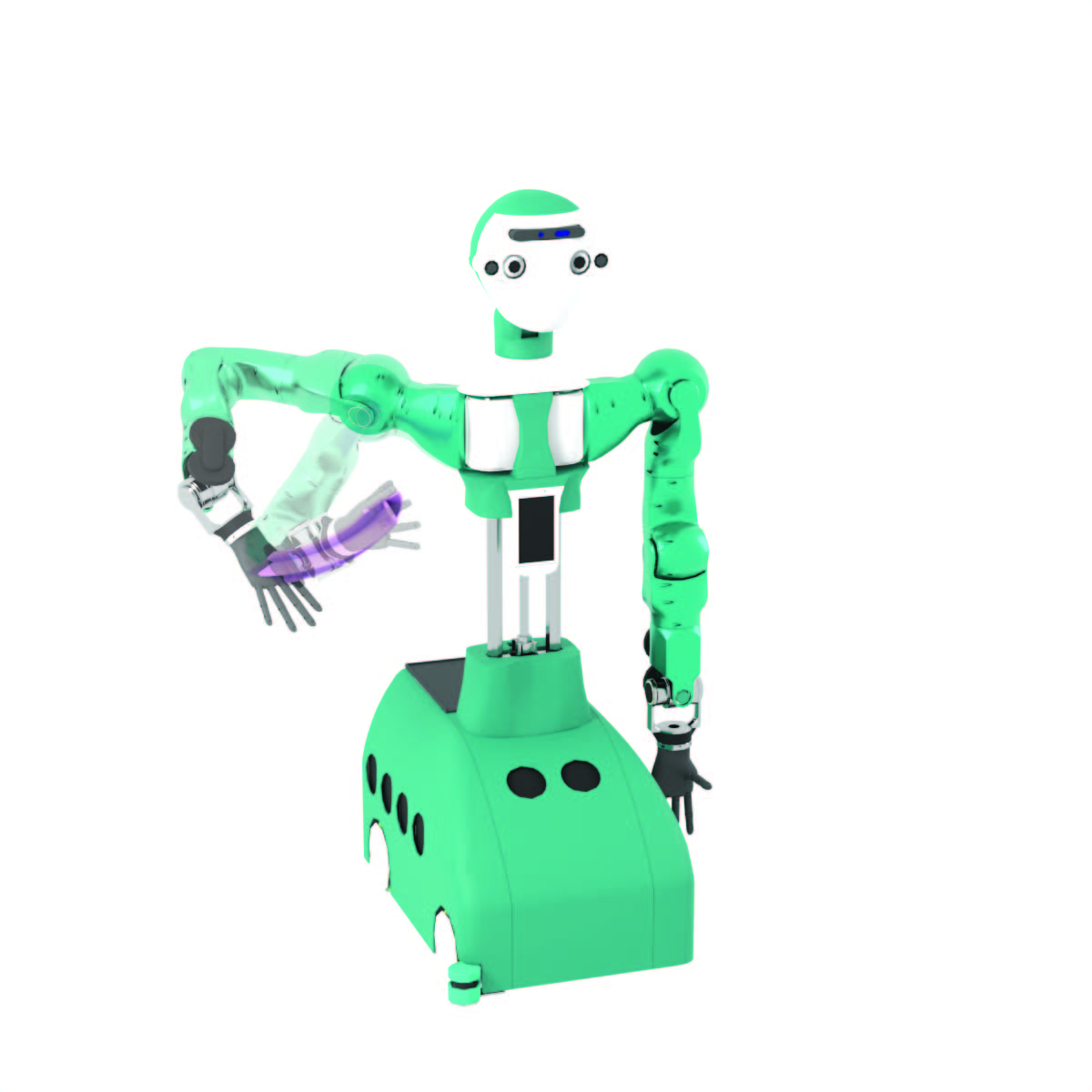}
		\caption{$w_1=-0.5$, $w_2=1.0$}
		\label{subFig:Combination3}
	\end{subfigure}
	\begin{subfigure}[b]{0.3\textwidth}
		\includegraphics[width=\textwidth,trim={5cm 17cm 10cm 5cm},clip]{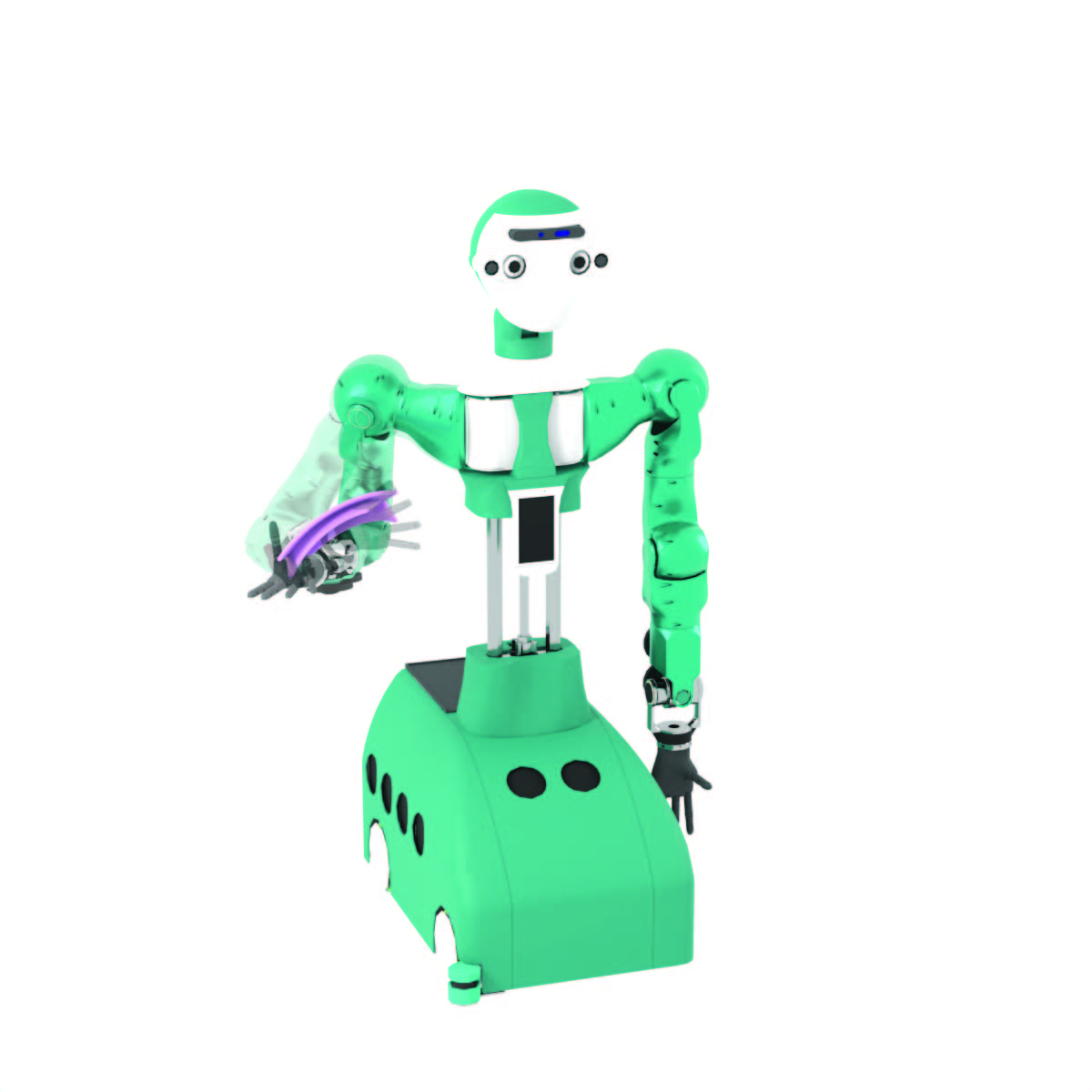}
		\caption{$w_1=-0.5$, $w_2=-0.5$}
		\label{subFig:Combination4}
	\end{subfigure}
	\caption{Illustrations of geodesic synergies for the right arm of the humanoid robot ARMAR-6~\cite{Asfour2019:Armar6}. \emph{(a)}-\emph{(b)} Arm motion (from transparent to opaque) and hand trajectory (\robotfirstGS, \robotsecondGS) resulting from individual synergies. \emph{(c)}-\emph{(f)} Arm motion and hand trajectory (\robotGScomb) obtained from different combinations of the two geodesic synergies \emph{(a)} and \emph{(b)}. } 	
	\label{Fig:ArmarExample}
	\vspace{-0.3cm}
\end{figure}
As previously mentioned, geodesics are the solution to the system of ODEs~\eqref{Eq:Geodesic} and are therefore completely determined by their initial conditions. In other words, one simply needs to define the initial joint configuration $\jointposition(0)\in\configmanifold$ and velocity $\jointvelocity(0)\in\configtangentspace{\jointposition(0)}$ and solve the corresponding initial value problem~\eqref{Eq:Geodesic} to obtain a minimum-energy trajectory in the robot configuration manifold. The resulting joint coordination is called a \emph{geodesic synergy}~\cite{Neilson15:GeodesicSynergyHypothesis}. 

Similarly to motion synthesis with classical synergies, novel behaviors can be obtained by combining several known geodesic synergies. However, their Riemannian nature must be taken into account, so that the combinations still result in meaningful, minimum-energy joint trajectories. 
Let us first consider two geodesic synergies with the same initial position $\jointposition_0$ and different initial velocities $\jointvelocity_0^{(1)}, \jointvelocity_0^{(2)}\in\configtangentspace{\jointposition_0}$. In this case, novel motions can be obtained by solving~\eqref{Eq:Geodesic} with the initial velocity defined as a weighted sum of the velocities of the individual synergies, i.e., $\jointvelocity_0=w_1 \jointvelocity_0^{(1)} + w_2 \jointvelocity_0^{(2)}$. However, geodesic synergies may not share a common initial position and it may be desirable that the robot starts its motion at a different initial configuration. In these cases, an additional step must be taken before summing the initial velocities. Indeed, as explained in Section~\ref{subSec:GeometryMechSys}, in Riemannian geometry, each velocity vector $\jointvelocity\in\configtangentspace{\jointposition}$ is linked to a specific position $\jointposition$ as it lies on its tangent space $\configtangentspace{\jointposition}$. Therefore, in order to be combined, the velocities $\jointvelocity_0^{(1)}\in\configtangentspace{\jointposition_0^{(1)}}$, $\jointvelocity_0^{(2)}\in\configtangentspace{\jointposition_0^{(2)}}$ must first be transported onto the tangent space of the initial configuration $\jointposition_0$. This is achieved via \emph{parallel transport}, a well-known operation in Riemannian geometry. Figure~\ref{Fig:ArmarExample} presents a proof of concept illustrating the generation of novel motions (Fig.~\ref{subFig:Combination1}-~\ref{subFig:Combination4}) via the combination of two geodesic synergies (Fig.~\ref{subFig:Geodesic1}-~\ref{subFig:Geodesic2}). It is important to notice that combinations of geodesic synergies are themselves geodesics. Moreover, the presented approach applies to any number of synergies. In particular, combinations of $N$ geodesic synergies with metric-orthogonal initial velocities (i.e., whose Riemannian inner product is $0$) can generate any motion of a $N$-DoFs robot. A video of the geodesic synergies of Figure~\ref{Fig:ArmarExample} including comparisons with traditional PCA-based linear synergies is available at \url{https://youtu.be/XblzcKRRITE}.

\vspace{-0.2cm}
\section{Vision and Challenges}
\vspace{-0.2cm}
\label{subSec:VisionAndChallenges}
As illustrated previously, geodesic synergies offer a compelling solution to generate energy-efficient, well-coordinated robot motions. Indeed, they inherit the benefits of the classical synergies, as a wide range of motions can simply be generated by combining few well-selected synergies. 
Moreover, they are specifically designed to account for the nonlinearities arising from the intrinsic dynamics of robotics systems. In this sense, geodesic synergies are physically meaningful, interpretable, and contribute to the explainability of the generated robot behavior. This contrasts with the PCA-based Euclidean synergies commonly used in robotics. Moreover, geodesic synergies directly encode dynamic robot motions, as opposed to postural and kinematic synergies, which disregard the robot dynamics.
Furthermore, geodesic synergies result in trajectories with minimum-energy demand for the robot. 
This becomes especially relevant for energy-constrained systems such as robotic prosthesis, exoskeletons, humanoid robots, and mobile robots in general. Finally, according to the recent studies~\cite{Biess07:ComputationalModelPointing,Neilson15:GeodesicSynergyHypothesis,Klein22:SequenceGeodesicSynergies} covered in Section~\ref{Sec:Motivation}, robot motions produced by geodesic synergies follow the same principle as in human motion planning. Therefore, they may lead to more natural motions of wearable robots that comply with human motion generation mechanisms and thus may improve their assistive abilities.
However, as discussed next, several challenges are yet to be tackled for leveraging geodesic synergies in robotics.

\vspace{-0.55cm}
\subsubsection{Selection of geodesic synergies}
Among the first challenges to solve for a successful application to robot motion generation is the extraction and selection of relevant geodesic synergies. To do so, we envision two different approaches. First, geodesic synergies may be learned from humans. Namely, we contend that geodesic synergies may be extracted from human motions using Riemannian dimensionality reduction methods such as principal geodesic analysis (PGA)~\cite{Fletcher04:PGA,Sommer14:ExactPGA}, the Riemannian equivalent of PCA. Similarly to PCA-based approaches~\cite{Gabiccini11:HandSynergiesGraspingForces,Gu06:HumanoidSynergies,Hauser07:KinematicSynergiesBalanceControl,Santello98:PosturalHandSynergies}, geodesic synergies may be given by the first $n$ principal geodesics accounting for a given proportion of the information contained in the motion. Notice that such an analysis may also offer novel perspectives on understanding the mechanisms of human motion generation. An important challenge would then be to transfer these biological geodesic synergies to robots, while accounting for the differences between the human and the robot configuration manifold. Although this may be relatively straightforward for robots with anthropomorphic design, this generally remains an open problem. A second approach to extract synergies would be to directly design them for the robot at hand. This may be achieved, e.g., by defining a orthonormal basis in the robot configuration manifold and select the most relevant directions (under some criteria) as geodesic synergies for the robot's movements. 

\vspace{-0.55cm}
\subsubsection{Interacting with the environment} Another challenge to be tackled in the context of geodesic synergies is to handle external influences arising from interactions with the environment or external perturbations. This is important as robot motions should not only remain optimal under all circumstances, but also be reliably safe for the user. We hypothesize that two research directions may be followed and coupled to tackle this challenge. First, the motion of the robot may be adapted to external conditions by artificially shaping the Riemannian metric, and which adapt the the geodesic synergies accordingly. To do so, we may take inspiration from methods developed around Riemannian motion policies~\cite{Cheng21:RMPflow,Ratliff18:RiemannianMotionPolicies}. This first approach may typically be used for the robot to cope with an additional load, e.g., an manipulated object. Second, we advocate for the design of Riemannian optimal controllers based on geodesic synergies, which would allow the robot to react appropriately to external events.

\vspace{-0.3cm}
\subsubsection{Geodesic synergies for action-perception couplings} To achieve motions adapted to a wide variety of tasks, the activation of synergies should be designed according to the current task and environment state. In other words, synergies, i.e., low-level robot actions, should be coupled to the perception. We hypothesize that geodesic synergies may be key components of novel, interpretable perception-action couplings in technical cognitive systems thanks to their Riemannian nature. Namely, taking inspiration from neurosciences~\cite{Neilson21:VisuallyGuidedMovements,Pellionisz85:FunctionalGeometriesCNS}, we suggest that robot perception spaces may be understood as geometric spaces and eventually endowed with Riemannian metrics, thus being identified as Riemannian manifolds. Through the lens of Riemannian geometry, perception inputs may then be intertwined with geodesic synergies via geometric mappings between their respective manifolds. Although important challenges such as the design of Riemannian perception spaces, and learning of mappings between the perception manifold and the configuration space manifold awaits along the way, we believe that introducing such geometric representations in robot perception-action loops may pave the way towards the generation of efficient and explainable robot motions. This may complement and improve upon recent end-to-end deep learning methods lacking explainability and adaptability.

Overall, we believe that Riemannian geometry may be the theory reconciling perception and action systems for robust robot motion generation. Viewing the robot action and perception spaces through the lens of geometry may lead to a unified Riemannian framework for robot motion learning, control, and adaptation. Such a framework would allow robots to seamlessly associate perception inputs with compatible and adaptable minimum-energy joint coordinations. We believe that this is key to provide robots with robust motion generation mechanisms that build on innovative and explainable perception-action loops. In this sense, Riemannian geometry may prove to be a game changer for deploying robots in our everyday life, as it was for Einstein's theory of general relativity.

\vspace{-0.3cm}
\subsubsection{\footnotesize Acknowledgements} \footnotesize This work was supported by the Carl Zeiss Foundation through the JuBot project. The authors also thank Leonel Rozo for his useful feedback on this paper and Andre Meixner for his help in rendering Figure~\ref{Fig:ArmarExample}.\normalsize

\vspace{-0.3cm}
\bibliographystyle{spmpsci}
\bibliography{References}  

\end{document}